# I2T2I: LEARNING TEXT TO IMAGE SYNTHESIS WITH TEXTUAL DATA AUGMENTATION

*Hao Dong, Jingqing Zhang, Douglas McIlwraith, Yike Guo*

Data Science Institute, Imperial College London

## ABSTRACT

Translating information between text and image is a fundamental problem in artificial intelligence that connects natural language processing and computer vision. In the past few years, performance in image caption generation has seen significant improvement through the adoption of recurrent neural networks (RNN). Meanwhile, text-to-image generation begun to generate plausible images using datasets of specific categories like birds and flowers. We've even seen image generation from multi-category datasets such as the Microsoft Common Objects in Context (MSCOCO) through the use of generative adversarial networks (GANs). Synthesizing objects with a complex shape, however, is still challenging. For example, animals and humans have many degrees of freedom, which means that they can take on many complex shapes. We propose a new training method called Image-Text-Image (I2T2I) which integrates text-to-image and image-to-text (image captioning) synthesis to improve the performance of text-to-image synthesis. We demonstrate that I2T2I can generate better multi-categories images using MSCOCO than the state-of-the-art. We also demonstrate that I2T2I can achieve transfer learning by using a pre-trained image captioning module to generate human images on the MPII Human Pose dataset (MHP) without using sentence annotation.

## 1. INTRODUCTION

As an image can be described in different ways and a caption can also be translated into different images, there is no bijection between images and captions. For example, in the sentence "a group of giraffes standing next to each other", the number of giraffes, the background and many other details are uncertain. It is an open issue of text-to-image mapping that the distribution of images conditioned on a sentence is highly multi-modal. In the past few years, we've witnessed a breakthrough in the application of recurrent neural networks (RNN) to generating textual descriptions conditioned on images [1, 2], with Xu *et al.* showing that the multi-modality problem can be decomposed sequentially [3]. However, the lack of datasets with diversity descriptions of images limits the performance of text-to-image synthesis on multi-categories dataset like MSCOCO [4]. Therefore, the problem of text-to-image synthesis is still far from being solved.

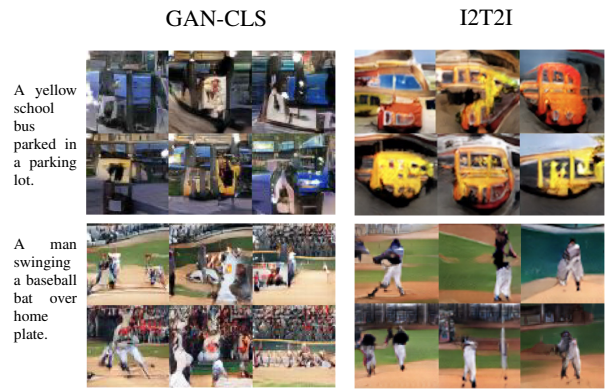

**Fig. 1**: Comparing text-to-image synthesis with and without textual data augmentation. I2T2I synthesizes the human pose and bus color better.

Recently, both fractionally-strided convolutional and convolutional neural networks (CNN) have shown promise for image generation. PixelCNN [5] learns a conditional image generator which is able to generate realistic scenes representing different objects and portraits of people. Generative adversarial networks (GAN) [6], specifically DCGAN [7] have been applied to a variety of datasets and shown plausible results. Odena *et al.* [8] has shown the capability of GANs to generate high-quality images conditioned on classes. To solve the multi-modality problem for text-to-image synthesis, Reed *et al.* [9] proposed GAN-CLS which bridges advances of DCGAN and an RNN encoder to generate images from an latent variables and embedding of image descriptions. Besides these, recent studies learned to generate images by conditions [10, 11]. However, GAN-CLS fails to generate plausible images on more complicated and changeable realistic scenes such as those illustrating human activities.

In this paper, we illustrate that sentence embedding should be able to capture details from various descriptions (one image can be described by plenty of sentences). This determines the robustness of understanding image detail. The main contribution of our work is a new training method called Image-Text-Image (I2T2I), which applies textual data augmentation to help text-to-image synthesis learn more descriptions of each image. I2T2I is composed of three modules: 1) the image captioning module 2) the image-text mapping module and 3) the GAN module. The image captioning module performs textual data augmentation and the other two modules are trained to realize text-to-image synthesis. Our method I2T2I achieves stronger capability on capturing details of descriptions. We evaluate our method by comparing images generated by GAN-CLS and I2T2I on the MSCOCO dataset. Furthermore, we demonstrate the flexibility and generality of I2T2I on MPII Human Pose dataset (MHP) [12] by using a pre-trained image captioning module from MSCOCO.

## 2. PRELIMINARIES

In this section, we briefly describe the previous work that our modules are based upon.

### 2.1. Image Captioning Module

Image captioning via deep CNN and RNN has seen significant improvements in recent years [2]. Deep CNNs can fully represent an image $X$ by embedding it into a fixed-length vector. Then RNN especially LSTM [13] decodes the fixed-length vector to a desired output sentence $S = (s_1, ..., s_t)$ by iterating the recurrence relation for $t$ defined in Equation (1) to (4):

$$h_0, c_0 = LSTM(\theta, W_e(CNN(X))) \quad (1)$$
$$h_t, c_t = LSTM(h_{t-1}, c_{t-1}, Ws(s_{t-1})) \quad (2)$$
$$p_t = softmax(h_t W_{ho} + b_o) \quad (3)$$
$$s_t = sample(p_t, k) \quad (4)$$

where $\theta$ is the initial hidden and cell states of LSTM which should be all zeros, $s_0$ is the special start token; $h_0$ is the hidden state initialized by the image embedding vector. Equation (2) sequentially decodes $h_0$ to $h_t$ in multiple steps, and Equation (3) gets the probabilities of words $p_t$. In every step, given $p_t$, the word token $s_t$ is sampled from top $k$ probabilities.

### 2.2. Image-Text Mapping Module

Visual-semantic embedding generates vector representation that embeds images and sentences into a common space [14]. A lot of work has been done to learn the multi-modal representation of images and sentences, where images and sentences can be mapped into the same embedding space by using CNN and RNN encoders [15]. The matching cost between images and sentences can be defined as Equation (5):

$$\min_\theta \sum_x \sum_k max\{0, \alpha - cs(x, v) + cs(x, v_k)\} + \sum_x \sum_k max\{0, \alpha - cs(x, v) + cs(x_k, v)\} \quad (5)$$

where $cs$ is cosine similarity of two embedded vectors, $x$ is embedded output of CNN encoder, $v$ is embedded output of RNN encoder. $x_k$ and $v_k$ are embedded outputs of mismatched images and sentences. The $\alpha$ is a margin value. This cost function maximizes the cosine similarity of mismatched pair of images and sentences, and minimizes the cosine similarity of matched images and sentences.

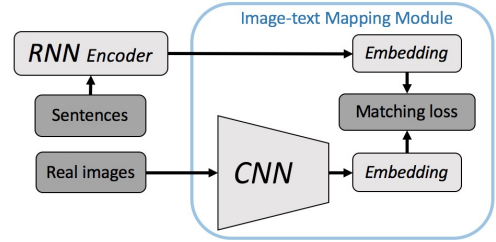

**Fig. 2**: Image and text mapping module

Figure 2 illustrates this approach to image-text mapping. The area marked in blue denotes image embedding and matching in a common latent space.

### 2.3. Generative adversarial network Module

Generative adversarial networks (GAN) are composed of a generator $G$ and a discriminator $D$, which are trained in a competitive manner [6]. $D$ learns to discriminate real images from fake ones generated by $G$, while $G$ learns to generate fake images from latent variables $z$ and tries to cheat $D$. The cost function is can be defined as Equation (6):

$$\min_G \max_D V(D,G) = E_{x \sim P_{sample}(x)}[log D(x)] + E_{z \sim P_z(z)}[log(1 - D(G(z)))] \quad (6)$$

Based on this, GAN-CLS [9] achieve text-to-image synthesis by concatenating text embedding features into latent variables of the generator $G$ and the last layer of discriminator $D$, as Fig. 3 shows.

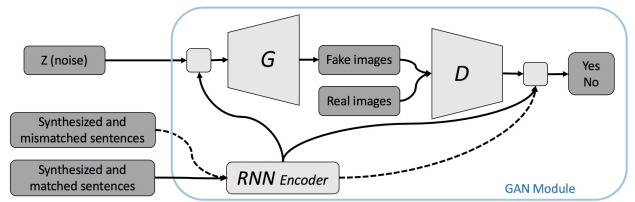

**Fig. 3**: GAN-CLS for text-to-image synthesis

Previously introduced GAN-CLS, instead of doing binary discrimination, the discriminator $D$ learns to discriminate three conditions, which are real images with matched text, fake images with arbitrary text, and real images with mismatched text. Figure 3 illustrates the architecture of a

GAN-CLS for text-to-image synthesis (in blue), and includes the previously introduced RNN encoder from image-text mapping module. Details of how we perform textual data augmentation – i.e. synthesize matched and mismatched sentences through our image captioning module, are given in Section 3.

## 3. METHOD

In this section, we introduce a significant contribution of our work – augmentation of textual data via image captioning. We demonstrate how this is performed through the use of a CNN and RNN, and how this is utilised in text-to-image synthesis.

### 3.1. Textual data augmentation via image captioning

For each real image $X_i$, there is a set $S_i$ that has a limited number of human-annotated captions, which cannot cover all possible descriptions of the images, as every sentence has lots of synonymous sentences. For example, "a man riding a surf board on a wave" is similar with "a person on a surf board riding a wave", "a surfer riding a large wave in the ocean" and etc. Fortunately, the softmax output of RNN-based image captioning can exploit the uncertainty to generate synonymous sentences. Therefore, with $k$ greater than 1 in Equation (4), a massive number of textual descriptions can be generated from a single image. Given an image $X$, the probability distribution of sentences $S$ can be defined as Equation (7):

$$P(S|X) = P(s_1|X)P(s_2|h_1, c_1, s_1)...$$
$$P(s_t|h_{t-1}, c_{t-1}, s_{t-1}) \quad (7)$$

Apart from synonymous sentences, the image captioning module may output similar but mismatched sentences. For example, "four birds are flying on the blue sky" and "five birds are flying in the sky".

Training the text encoder with massive synonymous sentences and similar sentences can help to improve the robustness of an RNN encoder. As Fig. 4 shows, with more synthetic sentences being used, an image will occupy more area in the text space and increase the coverage density.

Then, sentences not present in training datasets can be mapped to the nearest image, e.g, "five birds are flying in the sky" can be mapped with the image of "four birds are flying on the blue sky". The sentences that share semantic and syntactic properties are thus mapped to similar vector representations, e.g. "a man is riding a surfboard in a wave" can be mapped into "a man riding a surf board on a wave". As our method is able to map unseen or synonymous sentences to adjacent vector representation, improving the quality of image synthesis for these sentences.

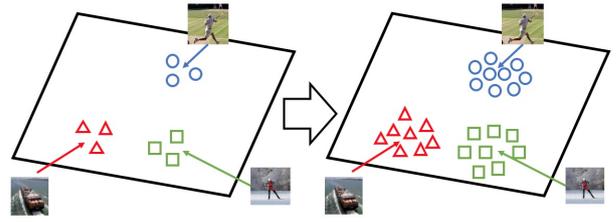

**Fig. 4**: Text and image space without (left) and with (right) textual data augmentation. After text augmentation the density of annotation is increased, providing a more robust training set.

### 3.2. Network structure

Our method includes three modules, previously introduced. These include 1) an image captioning module, 2) an image-text mapping module and 3) a conditional generative adversarial network for text-to-image synthesis (GAN-CLS). For our image captioning module, we use a state-of-the-art Inception model [16] followed by LSTM decoder as described in [1]. To improve the diversity of generated captions, we set $k = 3$ in Equation (4).

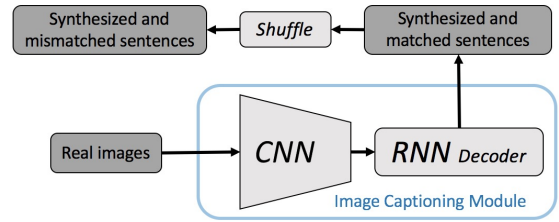

**Fig. 5**: Image captioning via convolutional neural network and recurrent neural network

To obtain image-text mapping (embedding), we follow the cost function defined in Equation (5), and use Inception V3 as the image encoder and LSTM as the sentence encoder. As image-text mapping requires both matched and mismatched sentences, after the image captioning module synthesizes a batch of matched sentences for a batch of images, we randomly shuffle and use them as mismatched sentences for image-text mapping module, as Fig. 5 shows.

We adopt GAN-CLS as image generation module, and reuse the LSTM encoder of image-to-text mapping module, then concatenate its embedded output into latent variables [9]. For the RNN encoder, we use LSTM [13] with a hidden size of 256. For GAN-CLS, in order to have a qualitative comparison, we use the same architecture describe in [9].

### 3.3. Training

Both matched and mismatched sentences are synthesized at the beginning of every iteration, then the GAN-CLS shown in

Fig. 3 will use these new sentences to learn image synthesis. As synthetic sentences are generated in every iteration, the RNN encoder for image and text embedding in our method is trained synchronously with the GAN-CLS module. Besides, in each iteration, we perform data augmentation for images includes random rotation, flip, and cropping.

For training the RNN encoder, we use learning rate of 0.0001, and the Adam optimization [17] with momentum of 0.5. For training GAN-CLS, we use initial learning rate of 0.0002 with Adam optimization with momentum of 0.5 also. Both learning rates of RNN encoder and GAN-CLS are decreased by 0.5 every 100 epochs. We use a batch size of 64 and train for 600 epochs.

## 4. EXPERIMENTS

Two datasets were used in our experiments. The MSCOCO [4] dataset contains 82783 training and 40504 validation images, each of which is annotated with 5 sentences from different annotators. The MPII Human Pose dataset (MHP) [12] has around 25K images covering 410 human activities. No sentence description is available, which makes it suitable to evaluate transfer learning.

### 4.1. Qualitative comparison

We evaluated our approach on the MSCOCO dataset with and without textual data augmentation applied. In both cases, the generator and discriminator use the same architectures i.e. GAN-CLS [9]. Results on the validation set can be compared in Fig. 1 and Fig. 6.

Both GAN-CLS and I2T2I generate a plausible background, but I2T2I performs better for certain classes such as "people" which has proved to be challenging for previous approaches [18]. Moreover, higher quality synthesis was found static object-based images such as window, toilet, bus etc.

### 4.2. Transfer learning

In order to demonstrate transfer learning, we used the pre-trained image captioning module from MSCOCO to train a text-to-image module on the MPII Human Pose dataset (MHP). As all images in this dataset contain people, we can successfully synthesize images of human activities, and the image quality is also better than GAN-CLS as Fig. 7 shows. This experiment shows that I2T2I can achieve text-to-image synthesis on unlabeled image dataset by using a pre-trained image captioning module from multi-categories dataset.

## 5. CONCLUSION

In this work, we propose a new training method I2T2I. The qualitative comparison results show that training text-to-image synthesis with textual data augmentation can help to obtain higher quality images. Furthermore, we demonstrate the synthesis of images relating to humans – which has proved difficult for existing methods [18]. In the immediate future, we plan to combine stackGAN [19] to improve image resolution.

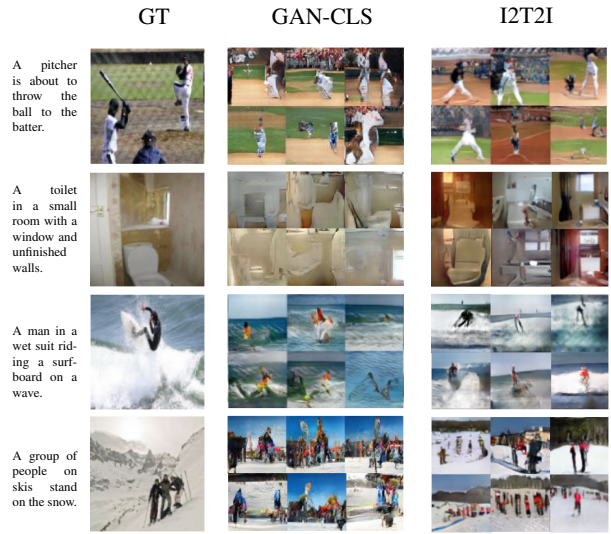

**Fig. 6**: Comparison between synthesized images using GAN-CLS and our I2T2I on the MSCOCO validation set. It is clear that I2T2I can better synthesize the human legs and waves compared with GAN-CLS. From the bathroom results, the I2T2I results have more detail on the window and toilet. Overall, I2T2I can better synthesize both simple stationary objects and those with a complex shape – such as humans.

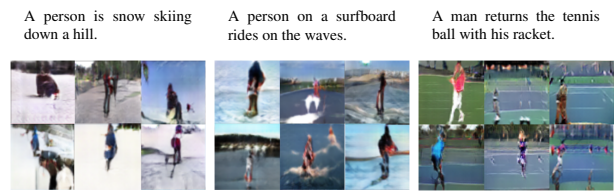

**Fig. 7**: Examples of synthesized images by transfer learning.

## 6. ACKNOWLEDGMENTS

The authors would like to thank Chao Wu and Simiao Yu for helpful comments and suggestions. Hao Dong is supported by the OPTIMISE Portal. Jingqing Zhang is supported by LexisNexis.